\documentclass[final]{cvpr}

\usepackage{times}
\usepackage{epsfig}
\usepackage{graphicx}
\usepackage{amsmath}
\usepackage{amssymb}
\usepackage{booktabs}
\usepackage{caption}
\usepackage[pagebackref=true,breaklinks=true,colorlinks,bookmarks=false]{hyperref}

\def\cvprPaperID{21} 
\def\confYear{CVPR 2021}

\begin{document}

\title{High-level camera-LiDAR fusion for 3D object detection with machine learning}

\author{  Gustavo A.~Salazar-Gomez
\thanks{Equal contribution}
\\  
  Department of Automatics and Electronics\\
  Universidad Autónoma de Occidente\\
  {\tt\small gustavo.salazar@uao.edu.co} \\
\and
Miguel A.~Saavedra-Ruiz
\footnotemark[1]
\\
  Department of Automatics and Electronics\\
  Universidad Autónoma de Occidente\\
{\tt\small miguel.saavedra@uao.edu.co} \\

\and
Victor A.~Romero-Cano\\
  Department of Automatics and Electronics\\
  Universidad Autónoma de Occidente\\
{\tt\small varomero@uao.edu.co} \\
}

\maketitle

\begin{abstract}
This paper tackles the 3D object detection problem, which is of vital importance for applications such as autonomous driving. Our framework uses a Machine Learning (ML) pipeline on a combination of monocular camera and LiDAR data to detect vehicles in the surrounding 3D space of a moving platform. It uses frustum region proposals generated by State-Of-The-Art (SOTA) 2D object detectors to segment LiDAR point clouds into point clusters which represent potentially individual objects. We evaluate the performance of classical ML algorithms as part of an holistic pipeline for estimating the parameters of 3D bounding boxes which surround the vehicles around the moving platform. Our results demonstrate an efficient and accurate inference on a validation set, achieving an overall accuracy of 87.1\%.
\end{abstract}

\section{Introduction}

Over the preceding years self-driving vehicles have received attention among the research community as a result of their potential of improving mobility, safety and reliability of transportation systems 
\cite{ref_1}. However, one of the core capabilities needed to unveil the complete potential of self-driving vehicles is the ability to perceive the objects surrounding it in the 3D space 
\cite{ref_2}.

3D object detection allows autonomous agents to estimate the relative pose of multiple objects neighbouring an ego-vehicle. Modern Deep Learning (DL) methods have been extensively wielded to address this issue. Some of the most common methods work directly over point clouds with convoluted deep neural network architectures 
\cite{ref_3, ref_4} or by creating a frustum region proposal, traditionally employing a RGB camera and a depth sensor 
\cite{ref_1, ref_5, ref_6, ref_7}. Notwithstanding the astonishing results presented by these models, their implementation is usually occluded by the vast need of computational resources required to deploy them 
\cite{ref_8}. Furthermore, substantial amounts of labeled datasets such as 
nuScenes \cite{ref_9} are needed to obtain acceptable accuracy levels.

\section{Research problem and motivation}

In this paper we present a framework to address the mentioned issues by combining SOTA deep learning algorithms for 2D detection with low-complexity, classical ML algorithms. Particularly, we show how these techniques can leverage the use of camera and LiDAR information to create a frustum region proposal 
\cite{ref_5} and deliver 3D object detections with few data samples in real-time. Classic ML algorithms have been exploited to resolve unsupervised learning problems like clustering sparse point clouds 
\cite{ref_10, ref_11} or supervised learning ones as pose parameter regression 
\cite{ref_12}. Nevertheless, there has not been significant research efforts towards the employment of these techniques into an high-level camera-LiDAR fusion for 3D object detection system.

The aforementioned considerations thoroughly motivate the conceptualization of this work. This paper is driven by the following research question: How to develop a ML pipeline to detect vehicles in the 3D space leveraging the utilization of mature 2D object detectors, using camera-LiDAR data, to estimate 3D bounding boxes in real-time for self-driving applications?

\begin{table*}[ht]

\centering

\begin{tabular}{ccccccccccc}
\toprule

\multicolumn{1}{c}{\begin{tabular}[c]{@{}c@{}} Set \end{tabular}}
&\multicolumn{1}{c}{\begin{tabular}[c]{@{}c@{}}x\end{tabular}}
& \multicolumn{1}{c}{\begin{tabular}[c]{@{}c@{}}y\end{tabular}}
& \multicolumn{1}{c}{\begin{tabular}[c]{@{}c@{}}z\end{tabular}}
& \multicolumn{1}{c}{\begin{tabular}[c]{@{}c@{}}\(\psi\)\end{tabular}} 
& \multicolumn{1}{c}{\begin{tabular}[c]{@{}c@{}}w\end{tabular}} 
& \multicolumn{1}{c}{\begin{tabular}[c]{@{}c@{}}l\end{tabular}} 
& \multicolumn{1}{c}{\begin{tabular}[c]{@{}c@{}}h\end{tabular}}
& \multicolumn{1}{c}{\begin{tabular}[c]{@{}c@{}}Avg.\end{tabular}}
& \multicolumn{1}{c}{\begin{tabular}[c]{@{}c@{}}Avg. 3D\end{tabular}}
& \multicolumn{1}{c}{\begin{tabular}[c]{@{}c@{}}Avg. BEV \end{tabular}}
\\ 
\midrule
Training    & \(98.8\) & \(98.2\) & \(99.9\) & \(78.0\) & \(96.8\) 
& \multicolumn{1}{c}{\begin{tabular}[c]{@{}c@{}}\(94.2\)\end{tabular}} 
& \(99.8\) & \(95.1\) 
& \multicolumn{1}{c}{\begin{tabular}[c]{@{}c@{}}\(62.0\)\end{tabular}}
& \multicolumn{1}{c}{\begin{tabular}[c]{@{}c@{}}\(68.4\)\end{tabular}}
\\ 
\midrule
Test        & \(96.6\) & \(97.8\) & \(95.4\) & \(55.7\) & \(80.7\) 
& \multicolumn{1}{c}{\begin{tabular}[c]{@{}c@{}}\(88.4\)\end{tabular}} 
& \(95.0\) & \(87.1\) 
& \multicolumn{1}{c}{\begin{tabular}[c]{@{}c@{}}\(42.7\)\end{tabular}}
& \multicolumn{1}{c}{\begin{tabular}[c]{@{}c@{}}\(47.8\)\end{tabular}}
\\ 
\bottomrule
\end{tabular}
\caption{Evaluation metrics of the proposed method.}
\label{tab:results}
\end{table*}

\section{Technical contribution}

In order to address this problem, we wield a set of different classic ML algorithms to estimate the 3D bounding box parameters of a given vehicle. Initially, we adopt a similar approach as the one proposed in 
\cite{ref_5}, where a frustum region proposal is assembled, taking advantage of SOTA 2D object detectors  
\cite{ref_13} reason by which we will only focus on the subsequent steps. Subsequently, the point cloud instance inside the frustum proposal is segmented using the DBSCAN 
\cite{ref_14} algorithm. Finally, a global feature representation encoding the relevant information of the given segmented instance 
is used as input for a Support Vector Regressor (SVR) 
\cite{ref_15} to estimate the 3D bounding box parameters. The final goal is to estimate the \(x, y, z \) centroid coordinates, the box dimensions \(w, l, h\) and its heading \(\psi\). 

\begin{figure}[h]
\begin{center}
   \includegraphics[width=0.9\linewidth, height = 0.45\textheight, keepaspectratio]{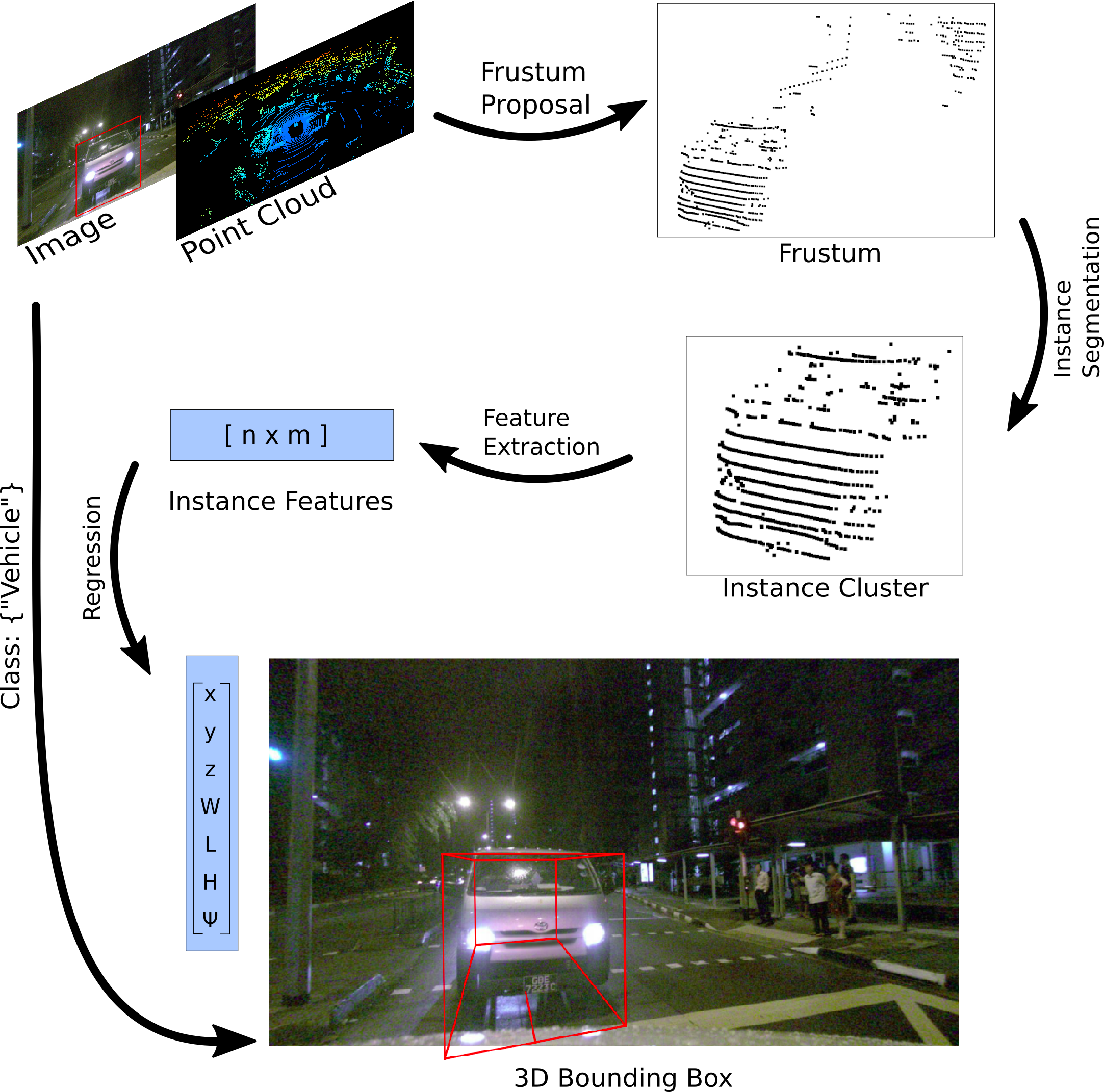}
\end{center}
   \caption{Proposed framework for 3D object detection.}
\label{fig:fig1}

\end{figure}

Other algorithms were tested like Random Forest \cite{ref_16}, XGBoost \cite{ref_17} for regression and KMeans \cite{ref_18} for clustering.  Over different experiments, the SVR algorithm exhibited the best results in terms of accuracy at a similar computational cost and less parameterization as the other regression algorithms. Likewise, KMeans constrained the clustering stage of the process as a specific number of cluster had to be set regardless that a scene may have different objects within.  For the sake of space, only the assessment of the pipeline with SVR and DBSCAN is presented in this work.

Our proposed framework is presented in Fig~\ref{fig:fig1} and it follows the steps previously mentioned. The nuScenes dataset is composed of 1000 scenes of 20 seconds each. In this work we employed a small version of this dataset called ``Mini" which contains a total of 10 scenes \cite{ref_9}. Our model was trained and tested using a total of \(1420\) image samples with the original image size provided in the dataset of \(1600\times900\) pixels. All the images used have at least one vehicle within and the dataset was split in \(80\%\) for training and \(20\%\) for testing. To assess performance, the 3D Intersection Over Union (IoU) was used to measure the percentage of volume intersected between the predicted bounding box and the ground truth. 

\begin{figure}[h]
\begin{center}
   \includegraphics[width=0.9\linewidth, height= 0.3\textheight, keepaspectratio]{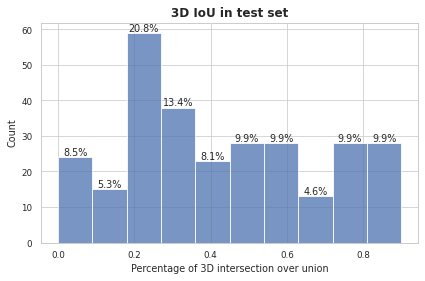}
\end{center}
   \caption{3D intersection over union results of ground truth versus predictions for the test set.}
\label{fig:fig2}
\end{figure}

\begin{table*}[ht]
\centering
\begin{tabular}{ccccc}
\toprule
Module          & \multicolumn{1}{c}{\begin{tabular}[c]{@{}c@{}}Instance \\ Segmentation\end{tabular}} & \multicolumn{1}{c}{\begin{tabular}[c]{@{}c@{}}Feature \\ Extraction\end{tabular}} & Regression & \textbf{Total} \\ \midrule
Training  &   \(11.1 s  \pm 2.3 s\)          &                       \(19.1 s  \pm 0.5 s\)       &
            \(2.6 s  \pm 8.3 ms\)          &          \( 32.8 s \pm 2.35 s\)        \\ \midrule
Inference &   \(4.7 ms \pm 4.8 ms\)          &                    \( 13 ms \pm 1.4 ms \)        &
                \(0.7 ms \pm 1.2 ms\)  & 
               \(18.4 ms \pm 5.1 ms\)                           \\
\bottomrule
\end{tabular}
\caption{Processing times through training and inference stages per module.}
\label{tab:stats}
\end{table*}

As shown in Fig~\ref{fig:fig2} approximately the \(44\%\) of predictions have an IoU above the \(50\%\). The thorough assessment of the 3D IoU and Birds-Eye-View (BEV) metrics can be seen in Table~\ref{tab:results}.

To obtain a thoroughly evaluation and establish which parameters are affecting the 3D IoU score, the accuracy of each one is presented in Table~\ref{tab:results}. From there it is possible to see that the proposed framework is capable of accurately estimating the centroid coordinates \(x, y, z\) and the bounding box dimensions \(w, l, h\) except for the width which presents an accuracy of approximately \(80\%\) in the test set. 
The low performance achieved in \(\psi\) is due to the difficulty estimating orientations with supervised learning techniques. 

Additionally, in Table~\ref{tab:results} is shown how with few data samples our proposed framework\footnote{\url{https://github.com/MikeS96/3d_obj_detection}} is capable to achieve an overall accuracy of \(87.1\%\) for the validation set with an average inference time of \(18.4ms\) per image and point cloud pair, using a \(3.20\) GHz CPU for training and inference. The aforementioned results validates how using a small number of samples from the original dataset, classical ML algorithms are capable to produce promising results with limited data and computational results. In fact, deep learning based SOTA algorithms are generally trained with huge datasets as the nuScenes which is composed of approximately 1.4 million of images and 390 thousand LiDAR sweeps in order to produce accurate results \cite{ref_9}.

To assess the processing time of our high-level architecture, Table~\ref{tab:stats} presents the training and inference time within each stage of the process. There it is possible to notice how our method is capable to train the whole system with 1136 images and LiDAR sweeps in roughly \(32.8s\) and process a new data sample in approximately \(18.4ms\) or 55FPS in a CPU-only setup. Compared with SOTA methods such like \cite{ref_19} which has 11FPS using a Titan RTX GPU, our method excels in processing time using limited amounts of data and computational resources.

\begin{table}[htb]
\centering
\begin{tabular}{@{}llllll@{}}
\toprule
Method   & Cam        & Rad        & LiDAR       & mASE & mAOE \\ \midrule
CenterFusion & \multicolumn{1}{c}{\checkmark} & \multicolumn{1}{c}{\checkmark} &   \multicolumn{1}{c}{-}          &  \textbf{0.142}    &  \multicolumn{1}{c}{\textbf{0.085}}   \\ \midrule
Ours     & \multicolumn{1}{c}{\checkmark} & \multicolumn{1}{c}{-} & \multicolumn{1}{c}{\checkmark}  &   0.573   & \multicolumn{1}{c}{0.840}  \\ \bottomrule
\end{tabular}
\caption{Performance comparison with Baseline for 3D object detection on nuScenes Dataset for the Car category.}
\label{tab:base}
\end{table}

We use the CenterFusion \cite{ref_20} model as baseline, in order to compare our results. This was submitted for nuScenes detection challenge in CVPR 20, and represents the SOTA of models that rely in frustum region to perform 3D object detection, what makes it the most suitable method to compare with. The authors stated that the model was trained using nuScenes dataset where two Nvidia P5000 GPUs were employed and the images size were reduced to \(800\times450\) pixels in order to increase computational speed. The evaluation metrics used in the comparison are; Average Scale Error (ASE) that is calculated as 1 – IoU after aligning centers and orientation, and Average Orientation Error (AOE) that is the smallest yaw angle difference between prediction and ground-truth in radians, where values close to zero are better. The results used for this assessment were the ones provided by the authors in the original paper \cite{ref_20}.

Table~\ref{tab:base} presents the results for the Car category. It shows whether the methods use radar or LiDAR for measuring depth. CenterFusion obtained 0.142 on ASE metric and compared with our proposed method which obtained an ASE of 0.573, we found that this difference may part from the robustness of the first one and enhancements such as Data Augmentation. Furthermore, our model is based on classical ML algorithms which makes the models less complex and faster to compute, necessary assets if working on scarce computational resources.

By comparing the AOE metrics we found a significant contrast with the baseline that shows 0.085 and ours that obtained 0.840, due to our proposed method does not use techniques such as MultiBin \cite{ref_21} architecture for orientation estimation, commonly used to solve this task.

Although the provided results do not surpass the current SOTA methods in 3D object detection in terms of accuracy, the given performance can be considered positive in scenarios where there are few amounts of data and the computational resources are limited. Additionally, comparing the results of our framework with the baseline, it is possible to notice that our models is capable to train and perform inference using only a CPU whereas the baseline relies on the use of expensive GPUs, reason by which our high-level framework could be deployed within low-cost settings. With improvements such as the implementation of MultiBin for heading estimation, the evaluation metrics could be considerable boosted toward better results.

\section{Conclusion}

In this work, we propose a framework that is capable of predicting 3D  bounding  boxes  for  vehicles  and  shows  promising  results estimating its parameters using classic ML techniques. Comparing our method with CenterFusion, it is possible to notice that even if the accuracy does not surpass the baseline\textquotesingle s, the performance in training and inference stages can be considered positive, and allows our framework to be deployed within low-cost settings and its use in real-time applications.


\medskip

\small


{\small
\bibliography{cvpr}
}

\end{document}